\documentclass{article}
\usepackage{ijcai19}

\usepackage{times}
\usepackage{soul}
\usepackage{url}
\usepackage[hidelinks]{hyperref}
\usepackage[utf8]{inputenc}
\usepackage[small]{caption}
\usepackage{graphicx}
\usepackage{amsmath}
\usepackage{booktabs}

\usepackage{amssymb}
\usepackage{amsmath}
\usepackage{amsthm}
\usepackage{subfig}
\usepackage{multirow}
\usepackage[linesnumbered,lined,ruled]{algorithm2e}

\SetCommentSty{mycommfont}

\newcommand{\nodes}{\mathcal{V}}

\newcommand{\validation}{\mathcal{Y}}

\newcommand{\loss}{\mathcal{L}}
\newcommand{\lossact}{\mathcal{L}_{\text{active}}}
\newcommand{\lossfro}{\mathcal{L}_{\text{frozen}}}
\newcommand{\fr}{\mathbb{R}}

\newcommand{\precision}{\text{Prec}}
\newcommand{\sgn}{\text{sgn}}

\title{Neurons Merging Layer: Towards Progressive \\ Redundancy Reduction for Deep Supervised Hashing}

\author{
Chaoyou Fu$^{1,2}$\footnote{Equal contribution. This work was done during the internship at Horizon Robotics.}\and
Liangchen Song$^{4*}$\and
Xiang Wu$^{1}$\and
Guoli Wang$^{4}$\And
Ran He$^{1,2,3}$\footnote{Ran He is the corresponding author.}\\
\affiliations
$^1$NLPR \& CRIPAC, Institute of Automation, Chinese Academy of Sciences\\
$^2$University of Chinese Academy of Sciences\\
$^3$Center for Excellence in Brain Science and Intelligence Technology, CAS\\
$^4$Horizon Robotics \\
\emails
\{chaoyou.fu, rhe\}@nlpr.ia.ac.cn,
alfredxiangwu@gmail.com,
\{liangchen.song, guoli.wang\}@horizon.ai
}

\begin{document}

\maketitle

\begin{abstract}
Deep supervised hashing has become an active topic in information retrieval. It generates hashing bits by the output neurons of a deep hashing network. During binary discretization, there often exists much redundancy between hashing bits that degenerates retrieval performance in terms of both storage and accuracy. This paper proposes a simple yet effective Neurons Merging Layer (NMLayer) for deep supervised hashing. A graph is constructed to represent the redundancy relationship between hashing bits that is used to guide the learning of a hashing network. Specifically, it is dynamically learned by a novel mechanism defined in our active and frozen phases. According to the learned relationship, the NMLayer merges the redundant neurons together to balance the importance of each output neuron. Moreover, multiple NMLayers are progressively trained for a deep hashing network to learn a more compact hashing code from a long redundant code. Extensive experiments on four datasets demonstrate that our proposed method outperforms state-of-the-art hashing methods.
\end{abstract}

\section{Introduction}
With the explosive growth of data, hashing has been one of the most efficient indexing techniques and drawn substantial attention \cite{lai2015simultaneous}.
Hashing aims to map high-dimensional data into a binary low-dimensional Hamming space.
Equipped with the binary representation, hashing can be performed with constant or sub-linear computation complexity, as well as the markedly reduced space complexity \cite{gong2011iterative}.
Traditionally, the binary hashing codes can be generated by random projection \cite{gionis1999similarity} or learned from data distribution \cite{gong2011iterative}.

Over the last few years, inspired by the remarkable success of deep learning, researchers have paid much attention to combining hashing with deep learning \cite{cao2016deep,zhu2017locality,lin2017discriminative,guo2017trivial,yang2018semantic,wu2018unsupervised}.
Particularly, by utilizing the similarity information for supervised learning, deep supervised hashing has greatly improved the performance of hashing retrieval \cite{li2015feature,jiang2017asymmetric}. 
In general, the last layer of a neural network is modified as the output layer of hashing bits.
Then, both features and hashing bits are learned from the neural network during optimizing the hashing loss function, which is elaborately designed to keep the similarities between the input data.
Convolutional neural network hashing (CNNH) \cite{xia2014supervised} is one of the early deep supervised hashing methods, which learns features and hashing codes in two separate stages.
On the contrary, deep pairwise-supervised hashing (DPSH) \cite{li2015feature} integrates the feature learning stage and hashing optimization stage in an end-to-end framework.
Recently, adversarial networks \cite{du2018redundancy,ma2018progressive,ghasedi2018unsupervised} and reinforcement learning \cite{yuan2018relaxation} are also applied to hashing learning.

\begin{figure}[t]
\centering
\includegraphics[height=6cm, width=0.45 \textwidth]{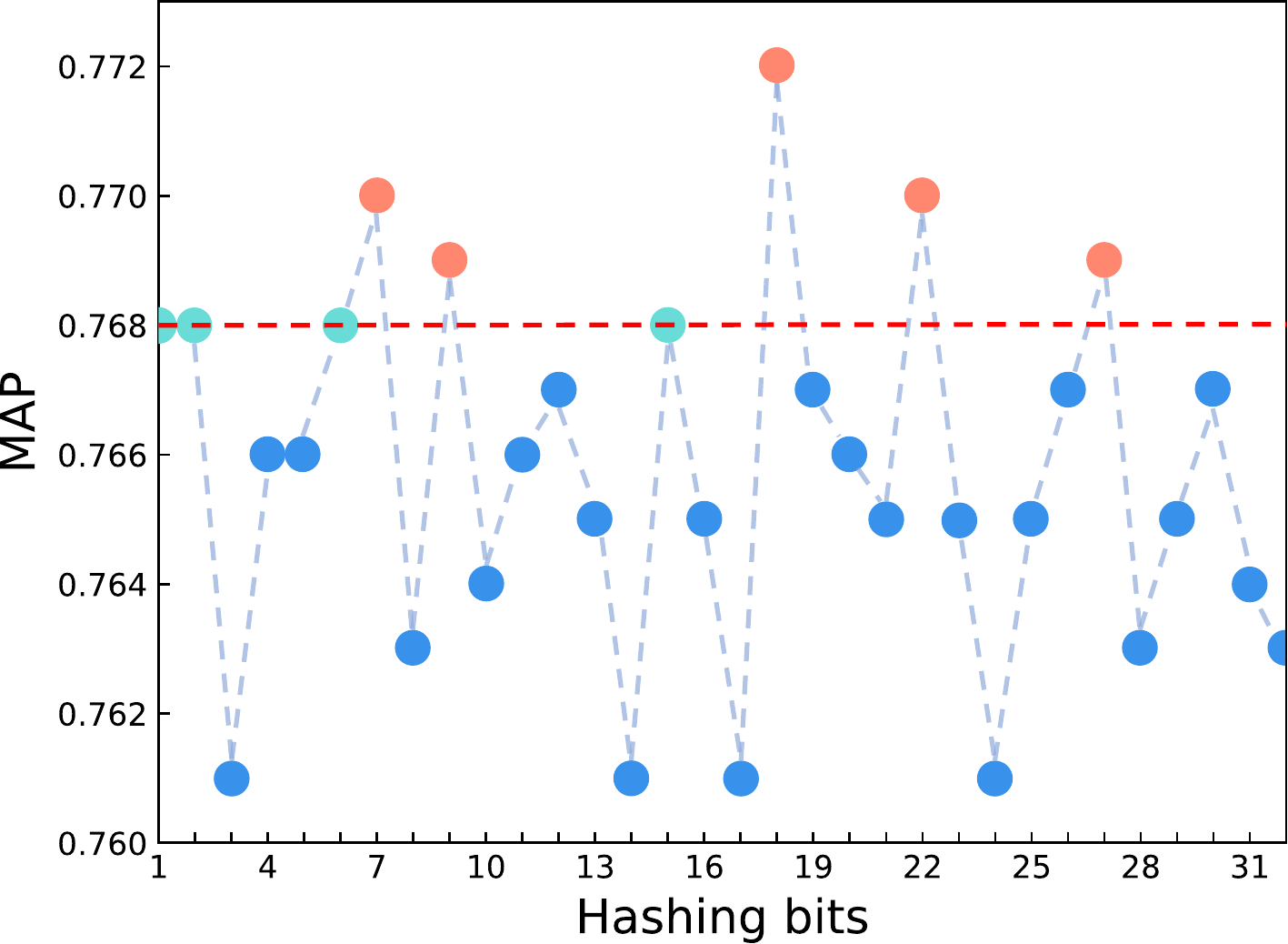}
\caption{Illustration of the redundancy in hashing bits generated by a common CNN-F network.
The horizontal red dotted line represents the Mean Average Precision (MAP) calculated using all bits. 
The vertical axis represents the MAP calculated after removing corresponding bit.
For example, removing the 1-st bit does not affect the MAP, while removing the 3-rd bit leads to a remarkable drop of MAP. Even more, the MAP increases after removing the 18-th bit. 
}
\label{fig:Redundancy_baseline}
\end{figure}

\begin{figure*}[t]
\centering
\includegraphics[width=0.95 \textwidth ]{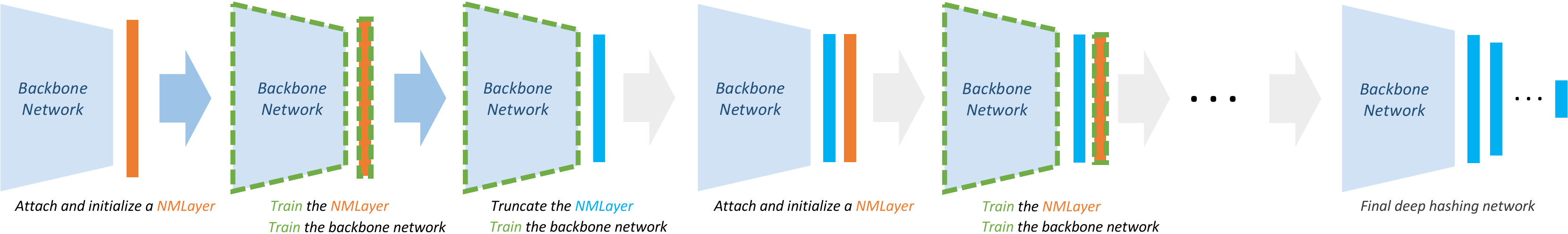}
\caption{Illustration of our progressive optimization strategy.
For a standard backbone network, i.e., deep hashing network, we first attach and initialize a NMLayer after the hashing layer.
Then, we train the NMLayer as well as the backbone network for a certain number of epochs.
Next, based on the learned adjacency relationship among neurons, the NMLayer is truncated to determine which neurons to merge.
After that, according to the truncation results, we continue to train the backbone network.
At this point, the training process of the first NMLayer is completed.
By iterating the above process, which is attaching NMLayer and optimizing the whole network, we finally get the required hashing bits.
Note that although many NMLayers are attached, total weights of the network change little.
Because each time completing the training of a NMLayer, we just determine which neurons in the hashing layer to merge, without adding extra weights.
}
\label{fig:Progressive}
\end{figure*}

Despite the effectiveness of the existing deep supervised hashing methods, the redundancy of hashing bits remains a problem that has not been well studied \cite{lai2015simultaneous,du2018redundancy}.
As shown in Figure~\ref{fig:Redundancy_baseline}, we can see that the redundancy has a significant impact on the retrieval performance.
Because of the redundancy, the importance of different hashing bits varies greatly.
However, a straightforward intuition is that all hashing bits should be equally important.
In order to address the redundancy problem, we propose a simple yet effective method to balance the importance of each bit in the hashing codes.
In details, we propose a new layer named Neurons Merging Layer (NMLayer) for deep hashing networks.
It constructs a graph to represent the adjacency relationship between different neurons.
During the training process, the NMLayer learns the relationship by a novel scheme defined in our active and frozen phases, as shown in Figure~\ref{fig:NMLayer}.
Through the learned relationship, the NMLayer dynamically merges the redundant neurons together to balance the importance of each neuron.
In addition, by training multiple NMLayers, we propose a progressive optimization strategy to gradually reduce the redundancy.
The full process of our progressive optimization strategy is illustrated in Figure~\ref{fig:Progressive}.
Extensive experimental results on the CIFAR-10, NUS-WIDE, MS-COCO and Clothing1M datasets verify the effectiveness of our method.
In short, our main contributions are summed up as follows:
\begin{enumerate}
	\item We construct a graph to represent the redundancy relationship between hashing bits, and propose a mechanism that consists of the active and frozen phases to effectively update the relationship. This graph results in a new layer named NMLayer, which reduces the redundancy of hashing bits by balancing the importance of each bit. The NMLayer can be easily integrated into a standard deep neural network.
	\item We design a progressive optimization strategy for training deep hashing networks. A deep hashing network is initialized with more hashing bits than the required bits, then the redundancy is progressively reduced by multiple NMLayers that form neurons merging. Compared with other hashing methods of fixed code length, NMLayers obtain a more compact code from a redundant long code.
	\item Extensive experimental results on four challenging datasets show that our proposed method achieves significant improvements especially on large-scale datasets, when compared with state-of-the-art hashing methods. 
\end{enumerate}

\section{Preliminaries and Notations}
\subsection{Notation}
We use uppercase letters like $A$ to denote matrices and use $a_{ij}$ to denote the $(i,j)$-th element in matrix $A$.
The transpose of $A$ is denoted by $A^{\top}$.
$\sgn(\cdot)$ is used to denote the element wise sign function, which returns $1$ if the element is positive and returns $-1$ otherwise.

\subsection{Problem Definition}
Suppose we have $n$ images denoted as $X = \{ x_{i} \}_{i=1}^{n}$, where $x_i$ denotes the $i$-th image.
Furthermore, the pairwise supervisory similarity is denoted as $S = \{s_{ij}\}$. 
$s_{ij} \in \{-1, +1\}$, where $s_{ij} = -1$ means $x_{i}$ and $x_{j}$ are dissimilar images and $s_{ij} = +1$ means $x_{i}$ and $x_{j}$ are similar images.

Deep supervised hashing aims at learning a binary code $b_i \in \{-1, +1\}^K$ for each image $x_i$, where $K$ is the length of binary codes.
$B = \{ b_{i} \}_{i=1}^{n}$ denotes the set of all hashing codes.
The Hamming distance of the learned binary codes of image $x_i$ and $x_j$ should keep consistent with the similarity attribute $s_{ij}$.
That is, similar images should have shorter Hamming distances, while dissimilar images should have longer Hamming distances.

\section{Neurons Merging Layer}
\begin{figure}[t]
	\centering
	\includegraphics[height=8.5 cm, width=0.45 \textwidth]{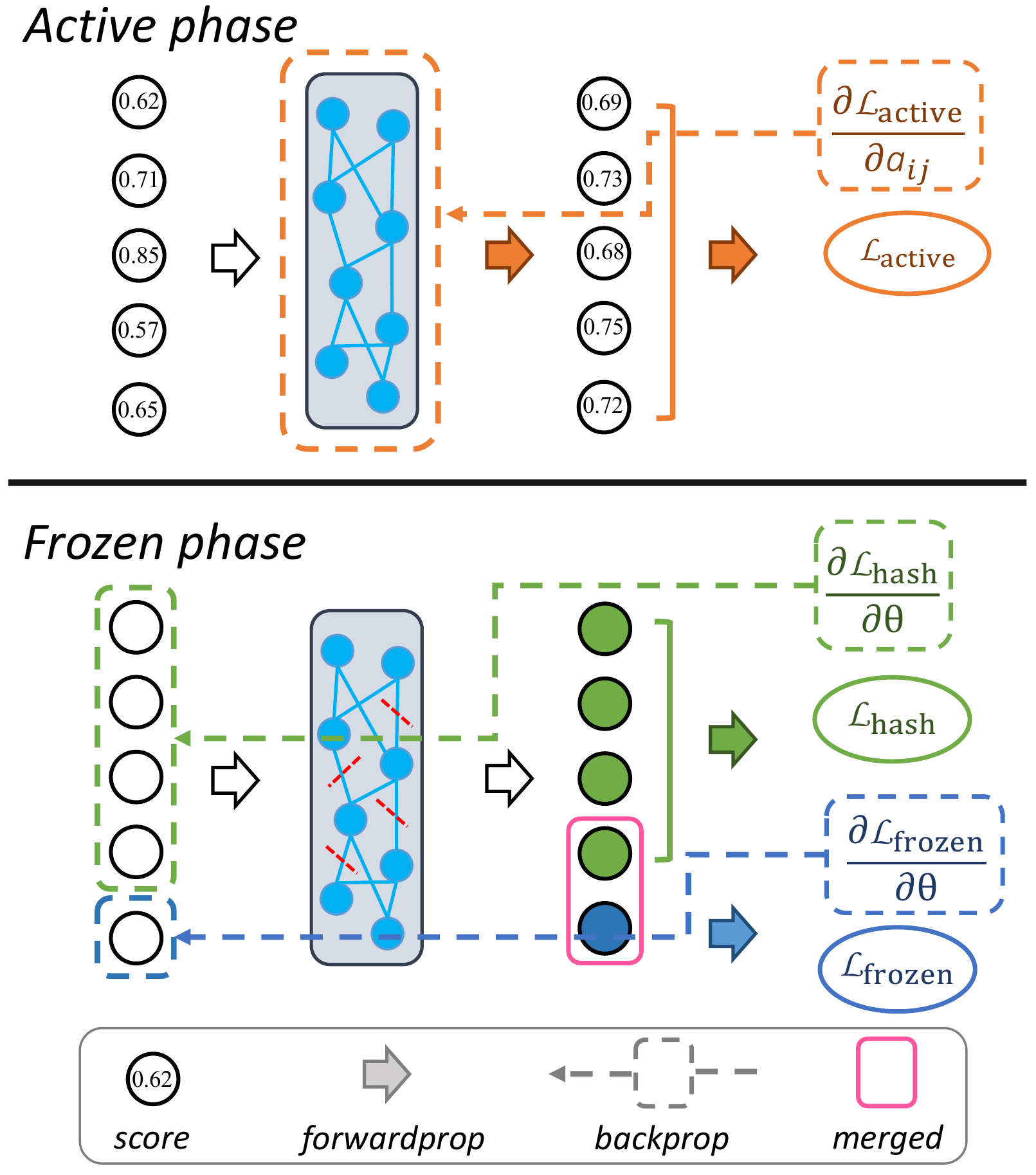}  
	\caption{Illustration of the two phases of our NMLayer. In the active phase, we calculate the score of each neuron and then utilize the active loss (Eq.~(\ref{eq:active})) to update the redundancy relationship, i.e., the adjacency matrix. In the frozen phase, after truncating the adjacency matrix, the hashing loss (Eq.~(\ref{eq:hash2})) and the frozen loss (Eq.~(\ref{eq:frozen})) are used to update corresponding neurons.}
	\label{fig:NMLayer}
\end{figure}

In this section, we describe the details of NMLayer, which aims at balancing the importance of each hashing output neuron.
A NMLayer has two phases during the training process, namely the active phase and the frozen phase, as shown in Figure~\ref{fig:NMLayer}. 
Basically, when a NMLayer is initially attached after a hashing output layer, it is set in the active phase to learn the redundancy relationship, i.e., the adjacency matrix, between different hashing bits.
After enough updating on the weights through backpropagation, we truncate the adjacency matrix to determine which neurons to merge.
Then, the NMLayer is set to frozen phase to learn hashing bits. 
The above process can be iterated for several times until the final output of the network reaches the required hashing bits.
Actually, the learning process of a NMLayer is constructing a graph $G$.
The neurons of a NMLayer are nodes, while the adjacency matrix $A$ denotes the set of all edges.
In the remainder of this section, we begin with presenting the basic structure of the NMLayer and then introduce the different policies of forward and backward in the active and frozen phases. 
Next, we define the behavior of the NMLayer when the neural network is in the evaluation mode.
Finally, we introduce our progressive optimization strategy in detail.
\subsection{Structure of the NMLayer}
As mentioned above, a NMLayer is basically a graph $G$ with learnable adjacency matrix $A$. 
Note that $G$ is an undirected graph, i.e., $a_{ij} = a_{ji}$.
The nodes of $G$ are denoted by $\nodes$, which is a set of hashing bits.
Specifically, the value type of $A$ differs in two phases.
During the active phase, $A$ is learned through backpropagation and $A\in\fr^{|\nodes|\times|\nodes|}$, where $|\nodes|$ means the number of nodes.
Each element $a_{ij}$ in $A$ denotes the necessity whether the two nodes $v_i$ and $v_j$ should be merged as one single node.
After entering frozen phase, the graph structure is fixed, that is $A$ becomes fixed and now $A\in\{0,1\}^{|\nodes|\times|\nodes|}$, where $a_{ij}=1$ means that the $i$-th and $j$-th neurons are merged, while $a_{ij}=0$ means the opposite.

\subsection{Active Phase}
When a NMLayer is first attached and initialized, all the elements in $A$ are set to 0, which indicates that no nodes are currently merged or inclined to be merged. 
In the active phase, our target is to find out which nodes should be merged together, based on a simple intuition that all nodes, i.e., all hashing bits, should carry equal information about the input data. 
In our NMLayer, the principle is restated in a practical way that eliminating any single hashing bit should lead to an equal decline of performance, thus no redundancy in the final hashing bits. 
Next, we elaborate on how to evaluate the importance of neurons in a typical forward pass of neural networks.

\paragraph{Forward.} Suppose the size of a mini-batch in a forward pass is $N$, the number of neurons is $K$, and the neurons are $\{v_1,\ldots,v_K\}$. 
In each forward pass, scores that evaluating the importance of each neuron are computed for the next backward pass. 
More precisely, for each neuron we compute the retrieval precision, i.e., Mean Average Precision (MAP), after eliminating it. 
We denote the input of the mini-batch as $\{X_n\}_{n=1}^N$ and the validation set as $\validation$, then the score $p_k$ of the $k$-th neuron is computed as
\begin{equation}
	p_k = \precision_k(X_n,\validation),
\end{equation}
where the function $\precision(X_n,\validation)$ means computing the precision with query $\validation$ and gallery $X_n$, and the subscript $k$ means computing precision without the $k$-th hashing bit.
Recall that in the active phase, elements in $A$ imply the necessity of whether two nodes in the graph should be merged.
In the forward pass, we take $A$ into consideration to calculate new scores $\{p_k'\}_{k=1}^K$, that is
\begin{equation}
	p_i' = p_i + \frac{1}{2} \sum_{i\neq j} a_{ij} (p_j - p_i).
\end{equation}
Next, we update $A$ according to the $\{p_k'\}_{k=1}^K$ in the following backward pass.

\paragraph{Backward.} In order to update $A$ through backpropagation, a loss function $\lossact$ is defined on $\{p_k'\}_{k=1}^K$. 
The principle of the loss function is to determine the inequality between neurons. 
Therefore, a feasible and straightforward loss function is 
\begin{equation}\label{eq:active}
	\lossact = \sum_{i\neq j} | p_i' - p_j' |.
\end{equation}
In fact, by Eq.~(\ref{eq:active}), the derivative of $\lossact$ with respect to $a_{ij}$ is
\begin{equation}\label{parital}
	\frac{\partial\lossact}{\partial a_{ij}} = \sgn(p_i' - p_j') \cdot (p_j - p_i).
\end{equation} 
Observing that the value of derivative depends on $p_j - p_i$.
It can be interpreted that the more different the two nodes are, the higher necessity the two nodes should be merged.

\subsection{Truncation of the Adjacency Matrix}
With $A$ being updated for several epochs, we then perform a truncation on $A$ to merge neurons.
After truncating $A$, all the elements in $A$ are either 0 or 1. 
Nodes with an adjacency value of 1 will be merged to reduce redundancy.
Note that, the strategy of truncation is various and we just use a straightforward one.
We turn the maximum $m$ values in $A$ to 1, and the others to 0. 
\vspace{0.2 cm}

\subsection{Frozen Phase}
If all the values in matrix $A$ are 0, that is the NMLayer neither trained nor truncated, both of the forward pass and backward pass are same as a normal deep supervised hashing network. 
When some elements in $A$ are 1, it means the corresponding nodes are merged together.
The new merged node that consists of several child nodes has new forward and backward strategies. 
Here, we illustrate our strategy with a simple case. 
Suppose that two nodes $\{v_1,v_2\}$ have been merged together after truncation, i.e., $a_{12}=1$.
Therefore, the length of the output hashing bits is now $K-1$, and we denote the new node as $v_{12}$, then the new output hashing bits are $\{v_{12},v_3,\ldots,v_K\}$.

\paragraph{Forward.} We randomly choose one child node from the new merged node as the output in the forward pass. In our simple example, suppose $v_1$ is randomly chosen, so the output of $v_{12}$ is equal to $v_1$.

\paragraph{Backward.} For the child node chosen as output in the forward pass, the gradient in the backward pass is simply calculated by the loss of hashing networks, such as a pairwise hashing loss like Eq.~(\ref{eq:hash2}).
As for those child nodes not chosen in the new merged node, we set a target according to the sign of the output of the chosen child node. 
In our simple example, the gradient of $v_1$ is calculated according to the pairwise hashing loss, while the gradient of $v_2$ is computed by $||v_2-\sgn(v_1)||^{2}$.
The intuition that not directly using the same gradient as $v_1$ is to reduce the correlation between the neurons.
More generally, for all of the child nodes in the new merged node expect $v_j$ chosen in the forward pass, the loss function is defined as
\begin{equation}\label{eq:frozen}
	\lossfro = \sum_{i\neq j} || v_i - \sgn(v_j) ||^{2}.
\end{equation} 

\subsection{Evaluation Mode}
When the whole network is set in evaluation mode, we no longer choose the output of a merged node in a random manner. 
Instead, we compute the output of the merged node by majority-voting. 
Again, using the simple example above, the output of $v_{12}$ depends on $\sgn(v_1)$ and $\sgn(v_2)$. 
That is, if $\sgn(v_1)=\sgn(v_2)=+1$, then $v_{12}=+1$. 
Note that when $\sgn(v_1)=+1$ and $\sgn(v_2)=-1$, then $v_{12}=0$, which implies that the output of $v_{12}$ is uncertain.
In this paper, we directly calculate the Hamming distance without considering this particular case and leave this study for our future pursuit. 

\subsection{Progressive Optimization Strategy}
By progressively training multiple NMLayers, we merge the output neurons of a deep hashing network as shown in Figure~\ref{fig:Progressive}.
It should be emphasized that in the training process, we only update the graph in a limited number of iterations.
In addition, during evaluation, the graph is fixed and no more calculations are required.
Therefore, the calculation of graph has little influence on the running time of the whole algorithm.
Note that we use multiple NMLayers instead of one because merging too many neurons at once will degrade algorithm performance,  which is reported in Figure~\ref{fig:Parameters}.
By performing the algorithm, we aim to get a network with $B_{\text{out}}$ hashing bits from a backbone network $F$ with $B_{\text{in}}$ hashing bits.
Hyper-parameters in the algorithm are shown as follow: $m$ means turning the maximum $m$ values of the adjacency matrix to $1$ and the others to $0$, which is defined in the truncation operation; 
the active phase and frozen phase are trained by $N_0$ and $N_1$ epochs respectively.

\section{Experiments}

\subsection{Experimental Details}

\paragraph{Pairwise Hashing Loss.} Following the optimization method in \cite{liu2012supervised}, we keep the similarity $s_{ij}$ between images $x_{i}$ and $x_{j}$ by optimizing the inner product of $b_{i}$ and $b_{j}$:
\begin{equation}
	\begin{aligned}
		\min \limits_{B} & \quad \loss_{\text{hash}} = \sum_{i= 1}^{m} \sum_{j= 1}^{n} (b_i^{\top} b_j - Ks_{ij})^2 \\
		\text{s.t.}      & \quad b_i,b_j \in \{ -1, +1\}^K
	\end{aligned}
	\label{eq:hash1}
\end{equation}
where $K$ denotes the length of hashing bits. $m$ and $n$ are the numbers of query images and retrieval images, respectively.
Obviously, the problem in Eq.~(\ref{eq:hash1}) is a discrete optimization problem, which is difficult to solve.
Note that for the input image $x_i$, the output of our neural network is denoted by $ u_i = F(x_i, \theta)$ ($\theta$ is the parameter of our neural network), and the binary hashing code $b_i$ is equal to $\sgn({u_i})$.
In order to solve the discrete optimization problem, we replace the binary $b_i$ with continuous $u_i$, and add a $L_{2}$ regularization term as \cite{li2015feature}. Then, the reformulated loss function can be written as
\begin{equation}
	\begin{aligned}
		\min \limits_{U, \Theta} & \quad \loss_{\text{hash}} = \sum_{i= 1}^{m} \sum_{j= 1}^{n}  (u_i^{\top} u_j - Ks_{ij})^2
		+ \eta \sum_{i = 1}^{n}\left \| b_i - u_i \right \|_2^2 \\
		\text{s.t.}              & \quad u_i,u_j \in \fr^{K\times 1}, b_{i} = \sgn(u_{i})
	\end{aligned}
	\label{eq:hash2}
\end{equation}
where $\eta$ is a hyper-parameter and Eq.~(\ref{eq:hash2}) is used as our basic pairwise hashing loss.

\paragraph{Parameter Settings.} In order to make a fair comparison with previous deep supervised hashing methods \cite{li2015feature,li2017deep,jiang2018deep}, we adopt CNN-F network \cite{chatfield2014return} pre-trained on ImageNet dataset \cite{russakovsky2015imagenet} as the backbone of our method.
The last fully connected layer of the CNN-F network is modified to hashing layer to output binary hashing bits.
The parameters in our algorithm are experimentally set as follows.
The number of neurons $B_{in}$ in hashing layer is set to 60.
In addition, the number of truncating edges in per step, i.e. $m$, is set to 4.
During training, we set the batch size to 128 and use Stochastic Gradient Descent (SGD) with $10^{-4}$ learning rate and $10^{-5}$ weight decay to optimize the backbone network.
Then, the learning rate of NMLayer and the hyper-parameter ${\eta}$ in Eq.~(\ref{eq:hash2}) are set to $10^{-2}$ and $1200$ respectively.
Moreover, the parameters $N_0$ and $N_1$ are set to 5 and 40 respectively.

\paragraph{Datasets.} We evaluate our method on four datasets, including \textbf{CIFAR-10} \cite{krizhevsky2009learning}, \textbf{NUS-WIDE} \cite{chua2009nus}, \textbf{MS-COCO} \cite{lin2014microsoft} and \textbf{Clothing1M} \cite{xiao2015learning}.
The division of CIFAR-10 and NUS-WIDE is the same with \cite{li2015feature}. The division of MS-COCO and Clothing1M is the same with \cite{jiang2017asymmetric} and \cite{jiang2018deep}, respectively. In addition, since a validation set is needed to calculate neuron scores in the active phase, we split the original training set into two parts: a new training set and a validation set. The number of validation set of CIFAR-10, NUS-WIDE, MS-COCO and Clothing1M is 200, 420, 400 and 280, respectively.

\paragraph{Evaluation Methodology.} We use Mean Average Precision (MAP) to evaluate retrieval performance.
For the single-label CIFAR-10 and Clothing1M datasets, images with the same label are considered to be similar ($s_{ij} = 1$).
For the multi-label NUS-WIDE and MS-COCO datasets, two images are considered to be similar ($s_{ij} = 1$) if they share at least one common label.
Specially, the MAP of the NUS-WIDE dataset is calculated based on the top 5,000 returned samples \cite{li2015feature,li2017deep}.

\begin{figure}[t]
	\centering
	\subfloat[]{
		\includegraphics[width=0.227\textwidth,height=3.5cm]{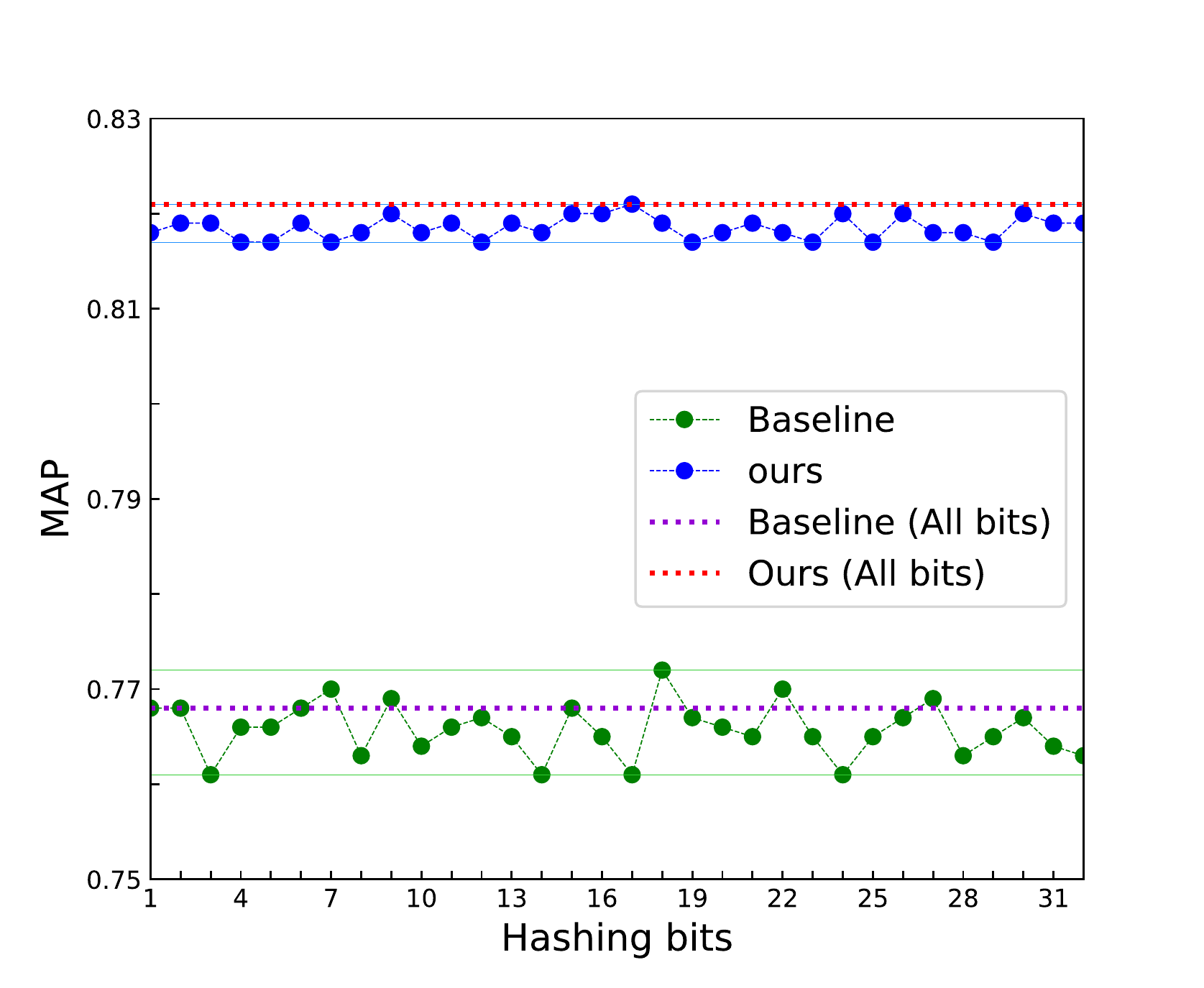} 
		\label{fig:Redundancy}
	}
	\subfloat[]{
		\includegraphics[width=0.227\textwidth,height=3.5cm]{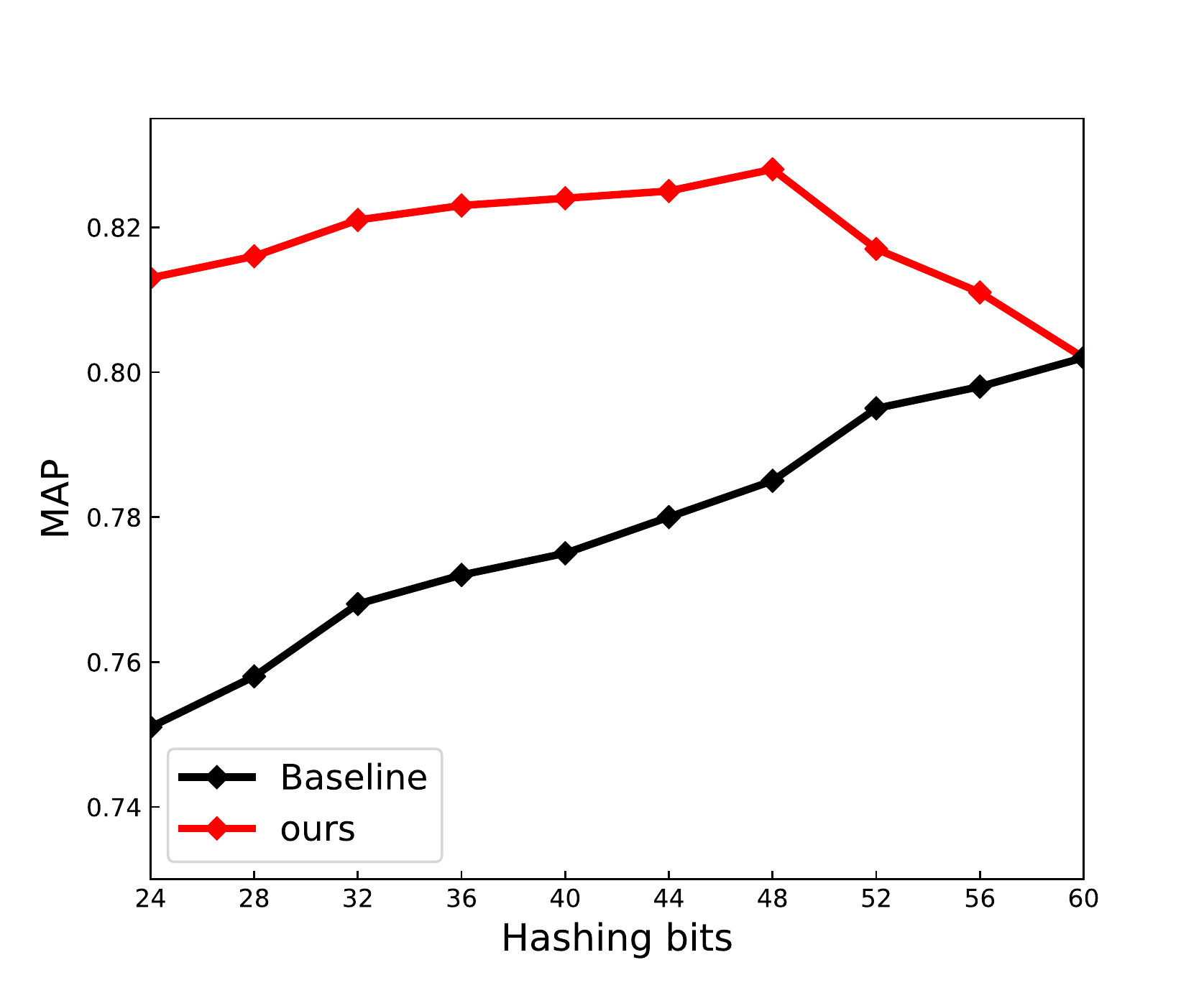}
		\label{fig:compression}
	}
	\caption{Analyses of redundancy in hashing bits. (a) Redundancy comparison between our method and baseline; (b) Bit reduction process. The red line denotes the MAP results during progressively reducing hashing bits from 60 to 24 (see from right to left). The black line denotes the MAP results of baseline when training the same fixed length hashing bits. }
	\label{fig:Redundancy_curves}
\end{figure}

\subsection{Experimental Results}
We compare our method with several state-of-the-art hashing methods, including one unsupervised method ITQ \cite{gong2011iterative}; four non-deep supervised methods, COSDISH \cite{kang2016column}, SDH \cite{shen2015supervised}, FastH \cite{lin2014fast} and LFH \cite{zhang2014supervised}; five deep supervised methods, DDSH \cite{jiang2018deep}, DSDH \cite{li2017deep}, DPSH \cite{li2015feature}, DSH \cite{liu2016deep}, and DHN \cite{zhu2016deep}. 

\setlength{\tabcolsep}{5pt} 
\begin{table*}[t]
	\begin{center}
		\resizebox{\textwidth}{!}{
		\begin{tabular}{|l|cccc|cccc|cccc|cccc|}
			\hline
			\multirow{2}{*}{Method}
			& \multicolumn{4}{|c}{CIFAR-10}
			& \multicolumn{4}{|c}{NUS-WIDE}
			& \multicolumn{4}{|c}{MS-COCO} 
			& \multicolumn{4}{|c|}{Clothing1M} \\
			\cline{2-5}
			\cline{6-9}
			\cline{10-13}
			\cline{14-17}
			        & 12 bits & 24 bits & 32 bits & 48 bits & 12 bits & 24 bits & 32 bits & 48 bits & 12 bits & 24 bits & 32 bits & 48 bits  & 12 bits & 24 bits & 32 bits & 48 bit \\
			\hline
			\textit{Ours}    &\bf 0.786  &\bf 0.813  & \bf 0.821 & \bf 0.828 & \bf 0.801 & \bf 0.824 & \bf 0.832 & \bf 0.840 & \bf 0.754   & \bf 0.772  & \bf 0.777 & \bf 0.782 & \bf 0.311 & \bf 0.372 & \bf 0.389 & \bf 0.401 \\
			\textit{Ours*}   & 0.750     & 0.797     & 0.813     & 0.825     & 0.774     & 0.812     & 0.827     & 0.832     & 0.744 & 0.769 & 0.775 & 0.780 & 0.268 & 0.343 & 0.377 & 0.396 \\
			\hline
			\hline
			DDSH    & 0.753     & 0.776     & 0.803     & 0.811     & 0.776     & 0.803     & 0.810     & 0.817     & 0.745 & 0.765 & 0.771 & 0.774 & 0.271 & 0.332 & 0.343 & 0.346 \\
			DSDH    & 0.740     & 0.774     & 0.792     & 0.813     & 0.774     & 0.801     & 0.813     & 0.819     & 0.743 & 0.762 & 0.765 & 0.769 & 0.278 & 0.302 & 0.311 & 0.319 \\
			DPSH    & 0.712     & 0.725     & 0.742     & 0.752     & 0.768     & 0.793     & 0.807     & 0.812     & 0.741 & 0.759 & 0.763 & 0.771 & 0.193 & 0.204 & 0.213 & 0.215 \\
			DSH     & 0.644     & 0.742     & 0.770     & 0.799     & 0.712     & 0.731     & 0.740     & 0.748     & 0.696 & 0.717 & 0.715 & 0.722 & 0.173 & 0.187 & 0.191 & 0.202 \\
            DHN     & 0.680     & 0.721     & 0.723     & 0.733     & 0.771     & 0.801     & 0.805     & 0.814     & 0.744 & 0.765 & 0.769 & 0.774 & 0.190 & 0.224 & 0.212 & 0.248 \\
			\hline
			COSDISH & 0.583     & 0.661     & 0.680     & 0.701     & 0.642     & 0.740     & 0.784     & 0.796     & 0.689 & 0.692 & 0.731 & 0.758 & 0.187 & 0.235 & 0.256 & 0.275 \\
			SDH     & 0.453     & 0.633     & 0.651     & 0.660     & 0.764     & 0.799     & 0.801     & 0.812     & 0.695 & 0.707 & 0.711 & 0.716 & 0.151 & 0.186 & 0.194 & 0.197 \\
			FastH   & 0.597     & 0.663     & 0.684     & 0.702     & 0.726     & 0.769     & 0.781     & 0.803     & 0.719 & 0.747 & 0.754 & 0.760 & 0.173 & 0.206 & 0.216 & 0.244 \\
			LFH     & 0.417     & 0.573     & 0.641     & 0.692     & 0.711     & 0.768     & 0.794     & 0.813     & 0.708 & 0.738 & 0.758 & 0.772 & 0.154 & 0.159 & 0.212 & 0.257 \\
			\hline
			ITQ     & 0.261     & 0.275     & 0.286     & 0.294     & 0.714     & 0.736     & 0.745     & 0.755     & 0.633 & 0.632 & 0.630 & 0.633 & 0.115 & 0.121 & 0.122 & 0.125\\
			\hline
			 
		\end{tabular}}
		\caption{MAP of different methods on CIFAR-10, NUS-WIDE, MS-COCO and Clothing1M datasets. \textit{Ours} denotes the results when $B_{in}$ is equal to 60, while \textit{Ours*} denotes the results when $B_{in}$ is equal to compared methods (12, 24, 32 and 48 respectively). Note that the MAP of NUS-WIDE dataset is calculated based on the top 5,000 returned samples.}
		\label{table:1}
	\end{center}
\end{table*}
\setlength{\tabcolsep}{5pt}

\begin{figure*}[t]
\centering
\includegraphics[width=0.85\textwidth]{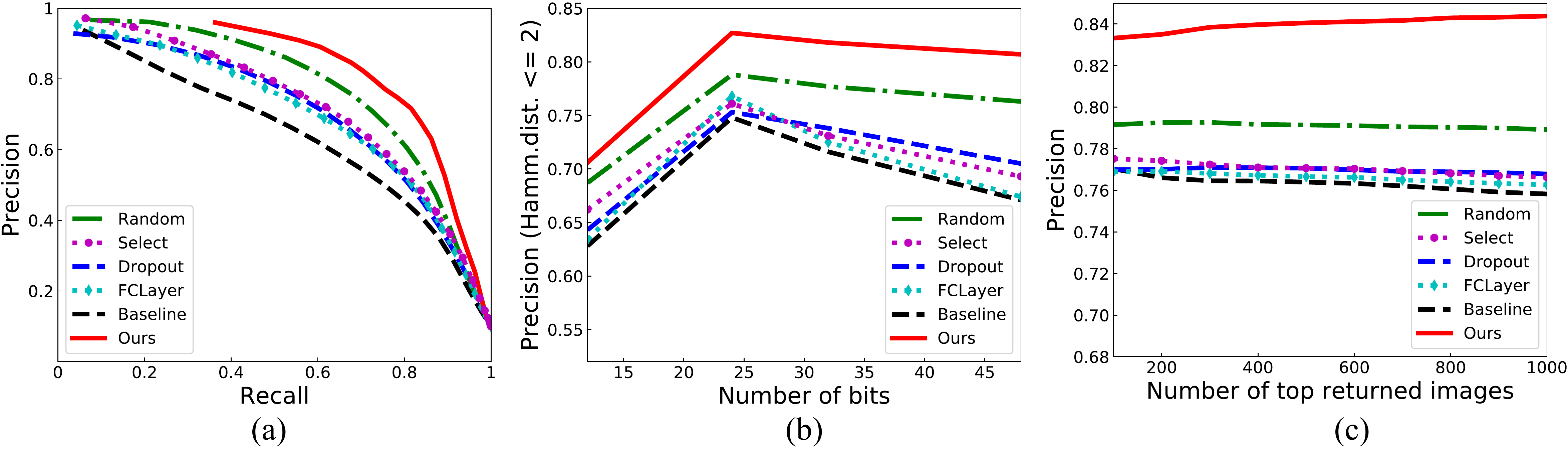}
\caption{The comparison results on CIFAR-10 dataset. (a) Precision-recall curves of Hamming ranking with 32 bits; (b) Precision curves within Hamming distance 2; (c) Precision curves with 32 bits w.r.t. different numbers of top returned samples.}
\label{fig:Metric_curves}
\end{figure*}

In Table~\ref{table:1}, the MAP results of all methods on CIFAR-10, NUS-WIDE, MS-COCO and Clothing1M datasets are reported.
For fair comparison, the results of DDSH, DSDH and DPSH come from rerunning the released codes under the same experimental setting, while other results are directly reported from previous works \cite{jiang2017asymmetric,jiang2018deep}.
As we can see from Table~\ref{table:1}, our method outperforms all other methods on all datasets, which validates the effectiveness of our method. It is obvious that our improvement is significant especially on large-scale Clothing1M dataset.

Besides, in Table~\ref{table:1}, we also report the results of the setting that $B_{in}$ is equal to compared hashing methods, which are denoted as \emph{ours*}.
Specifically, we set the number of initial neurons of hashing layer to 12, 24, 32 and 48 respectively, then the redundancy of these hashing bits is reduced by our method.
From Table~\ref{table:1}, we can observe that our method is superior to all other hashing methods with the setting of 24, 32 and 48 bits, except 12 bits.
The reason behind this phenomenon is that the redundancy of short code is essentially low.

\subsection{Experimental Analyses}

\subsubsection{Analyses of Redundancy in Hashing Bits}

On CIFAR-10 dataset, we train a 32-bits hashing network without NMLayer as a \emph{Baseline} based on Eq.~(\ref{eq:hash2}).
Then, in order to show the redundancy of hashing bits, we remove a bit per time and report the final MAP with our method and the baseline method in Figure~\ref{fig:Redundancy}.
It is clearly observed that compared to baseline, the variance of MAP of our algorithm is much lower, thus we can come to the conclusion that the redundancy in hashing bits has been reduced.
In addition, as the redundancy is reduced, each bit of hashing codes can be fully utilized.
Therefore, the retrieval precision of our method is greatly improved.

As shown in Figure~\ref{fig:compression}, compared with the baseline results trained on the fixed length hashing bits, we record the changes of MAP during progressively reducing hashing bits from 60 to 24.
As we can see from Figure~\ref{fig:compression}, the MAP value of our method increases from 60 to 48 bits.
At the same time, the curve of our method is more stable, while the baseline curve drops rapidly.
Both of these phenomenons are due to the effective redundancy reduction of our approach.
Finally, the MAP curve of our method reaches its maximum value at 48 bits.
Therefore, we consider 48 as the most appropriate code length on CIFAR-10 dataset.
Inspired by this insight, our approach can also be conducive to finding the most appropriate code length while reducing the redundancy.

\subsubsection{Comparisons with Other Variants}

In order to further verify the effectiveness of our method, we elaborately design several variants of our method.
Firstly, \emph{Random} is a variant of our method without active phase.
It replaces the dynamic learning adjacency matrix in the active phase with a random matrix.
Secondly, \emph{Select} is a variant of our method without frozen phase.
It directly selects the most important bits as the final output instead of merging them.
Thirdly, considering that the dropout technique \cite{srivastava2014dropout} is widely adopted in neural networks to reduce the correlation between neurons, we add a dropout layer before the hashing layer to reduce the correlation of hashing bits and denote it as \emph{Dropout}.
Finally, since the process of our neurons merging can be viewed as a process of dimension reduction, we design a variant \emph{FCLayer} to compare the differences between our NMLayer and the fully connected layer.
It replaces the NMLayer with a fully connected layer, which is optimized by loss function Eq.~(\ref{eq:hash2}).

The above variants are compared using three widely used evaluation metrics as \cite{xia2014supervised}: Precision curves within Hamming distance 2, Precision-recall curves and Precision curves with different numbers of top returned samples.
The results of above variants are reported in Figure~\ref{fig:Metric_curves}.
From Figure~\ref{fig:Metric_curves} we can see that compared to our method, the performance of both \emph{Random} and \emph{Select} has declined.
It demonstrates the validity of our active and frozen phases.
In addition, the improvements of \emph{Dropout} and \emph{FCLayer} over \emph{Baseline} are small, which proves the effects of the dropout technique and the fully connected layer are limited to the hashing retrieval.

\subsubsection{Sensitivity to Parameters}

\begin{figure}[t]
\centering
\includegraphics[width=0.45\textwidth]{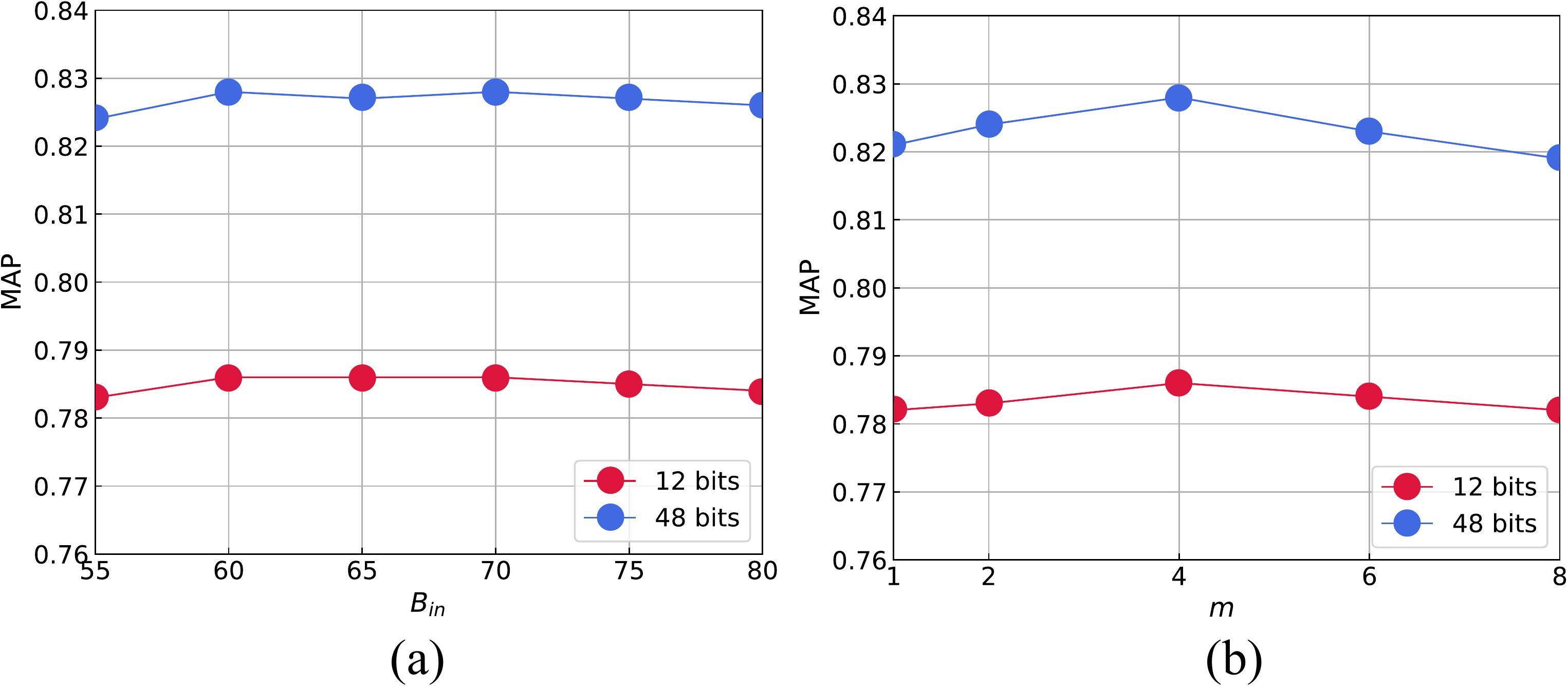}
\caption{Sensitivity study on CIFAR-10 dataset. (a) Sensitivity to parameter $B_{in}$; (b) Sensitivity to parameter $m$.}
\label{fig:Parameters}
\end{figure}

Figure~\ref{fig:Parameters} presents the effects of hyper-parameters $B_{\text{in}}$ and $m$. We can see that that increasing the number of $B_{\text{in}}$ dose not obviously improve the retrieval accuracy. It is due to that 60 bits already have enough expression capacity and extra neurons are saturated. Moreover, the retrieval results decrease when $m$ is too large, which demonstrates that merging too many neurons at once will degrade the performance of our algorithm. It also explains the necessity of our progressive optimization strategy.

\section{Conclusion}
In this paper, we analyze the redundancy of hashing bits in deep supervised hashing.
To address this, we construct a graph to represent the redundancy relationship and propose a novel layer named NMLayer.
The NMLayer merges the redundant neurons together to balance the importance of each hashing bit.
Moreover, based on the NMLayer, we propose a progressive optimization strategy.
A deep hashing network is initialized with more hashing bits than the required bits, and then multiple NMLayers are progressively trained to learn a more compact hashing code from a redundant long code.
Our improvement is significant especially on large-scale datasets, which is verified by comprehensive experimental results.

\section*{Acknowledgements}
This work is partially funded by National Natural Science Foundation of China (Grant No. 61622310), Beijing Natural Science Foundation (Grant No. JQ18017), Youth Innovation Promotion Association CAS(Grant No. 2015190).

{
\bibliographystyle{named}
\bibliography{ijcai19}
}

\end{document}